\documentclass[conference]{IEEEtran}
\usepackage{enumerate}
\usepackage{graphicx}
\usepackage{epsf}
\usepackage{subfigure}
\usepackage{amsmath}
\usepackage{amssymb}
\usepackage{array}
\usepackage{setspace}
\usepackage[amsmath,thmmarks,amsthm]{ntheorem}
\usepackage[ruled,lined]{algorithm2e}
\usepackage{algorithmic}
\usepackage{color}
\usepackage{amsfonts}
\usepackage{dsfont}
\usepackage{xfrac}
\usepackage{breqn}
\usepackage{bbm}
\usepackage{cite}
\usepackage{cases}
\usepackage{subeqnarray}

\graphicspath{{fig/}}

\theoremseparator{.}

\begin{document}
\title{\huge Intelligent Load Balancing and Resource Allocation in O-RAN: A Multi-Agent Multi-Armed Bandit Approach} 


\author{\IEEEauthorblockN{Chia-Hsiang Lai, Li-Hsiang Shen and Kai-Ten Feng}\\
 \IEEEauthorblockA{Department of Electronics and Electrical Engineering \\
  National Yang Ming Chiao Tung University, Hsinchu, Taiwan\\
Email: laivb22.ee10@nycu.edu.tw, gp3xu4vu6.cm04g@nctu.edu.tw, ktfeng@nycu.edu.tw}}

\maketitle

\begin{abstract}
The open radio access network (O-RAN) architecture offers a cost-effective and scalable solution for internet service providers to optimize their networks using machine learning algorithms. The architecture's open interfaces enable network function virtualization, with the O-RAN serving as the primary communication device for users. However, the limited frequency resources and information explosion make it difficult to achieve an optimal network experience without effective traffic control or resource allocation. To address this, we consider mobility-aware load balancing to evenly distribute loads across the network, preventing network congestion and user outages caused by excessive load concentration on open radio unit (O-RU) governed by a single open distributed unit (O-DU). We have proposed a multi-agent multi-armed bandit for load balancing and resource allocation (mmLBRA) scheme, designed to both achieve load balancing and improve the effective sum-rate performance of the O-RAN network. We also present the mmLBRA-LB and mmLBRA-RA sub-schemes that can operate independently in non-realtime RAN intelligent controller (Non-RT RIC) and near-RT RIC, respectively, providing a solution with moderate loads and high-rate in O-RUs. Simulation results show that the proposed mmLBRA scheme significantly increases the effective network sum-rate while achieving better load balancing across O-RUs compared to rule-based and other existing heuristic methods in open literature.
\end{abstract}
\begin{IEEEkeywords}
O-RAN, Near-RT RIC, Non-RT RIC,  O-DU, O-RU, load balancing, resource allocation, multi-agent, multi-armed bandit
\end{IEEEkeywords}

\section{Introduction}\label{INT}
Open radio access network (O-RAN) is the novel architecture proposed by O-RAN alliance for next-generation networks \cite{ORAN_1}. Unlike traditional base stations (BSs), the functions of an O-RAN BS are split among open central units (O-CUs), open distributed units (O-DUs), and open radio units (O-RUs) operating through software \cite{ORAN_2}. O-RAN alliance has also defined new network functions, non-realtime radio access network (RAN) intelligent controller (Non-RT RIC) and Near-RT RIC, which enable Internet service providers (ISPs) to deploy their own artificial intelligence (AI) based algorithms on these two functions \cite{ORAN_3}. Non-RT RIC is typically used as a place to train AI models in rApps, while Near-RT RIC executes the trained models for real-time network control and optimization in xApps.

As the number of mobile devices and emerging entertainment industries grows, users rely increasingly on the internet. Maintaining quality of service (QoS) for users is the primary concern for ISPs, but it is not an easy task. A large crowd of people flooding into an area can cause a BS to be overloaded, leading to network congestion. To avoid this, mobility load balancing (MLB) is necessary to share network traffic among neighboring BSs, allowing more efficient allocation of network resources to guarantee QoS for users. MLB has been the subject of recent research. \cite{MLB1} realizes energy conservation and QoS satisfaction by the proposed load-balanced user association and resource allocation algorithm. \cite{MLB2} focuses on WiFi networks, using average signal strength, channel occupancy, and access point load to make optimal handover decisions. For space-ground integrated networks, \cite{MLB3} uses a two-stage offloading mechanism to balance loads, increase network throughput, and satisfy more users. In \cite{MLB4} and \cite{MLB5}, load distribution is improved by adjusting BS parameters. \cite{MLB4} adjusts handover margin and time-to-trigger based on user speed and neighboring BS capacity. \cite{MLB5} predicts user positions using Bayesian additive regression trees and forecasts cell loads to calculate the cell individual offset (CIO) value.

In addition to non-AI-based approaches, using AI and machine learning (AI/ML) to solve problems is becoming increasingly popular in various fields due to advances in computing power. Recent studies on network management problems have utilized deep learning and reinforcement learning (RL). For example, \cite{AIMLB1, AIMLB2, AIMLB3, AIMLB4, AIMLB5, AIMLB6, AIMLB7} tune the CIO of BSs using RL algorithms to achieve load balancing in networks. Q-learning is used in \cite{AIMLB1} and \cite{AIMLB2}, while deep RL (DRL) is used in \cite{AIMLB3, AIMLB4, AIMLB5, AIMLB6, AIMLB7}. \cite{AIMLB3} and \cite{AIMLB4} view the problem as a multi-agent problem and use the deep deterministic policy gradient (DDPG) algorithm to solve it. \cite{AIMLB5} and \cite{AIMLB6} consider not only adjusting the CIO value but also the transmit power of the BSs, with the former using the twin delayed DDPG approach and the latter applying a dueling double Q-network algorithm. \cite{AIMLB7} proposes the conservative deep Q-learning algorithm to address Q-value overestimation problems in \cite{AIMLB6}. Other studies, such as \cite{RA1,RA2,DRL1,DRL2,QLEARNING1,QLEARNING2,MAB1,MAB2,MAB3}, focus on solving radio resource management problems. Non-AI-based approach is adopted in \cite{RA1,RA2}, DRL is used in \cite{DRL1,DRL2}, Q-learning is applied in \cite{QLEARNING1,QLEARNING2}, and the multi-armed bandit (MAB) method is implemented in \cite{MAB1,MAB2,MAB3}. The MAB method differs from other RL approaches in that the learning framework only includes action and reward, effectively reducing the dimension of the action searching space and the computation overhead of the algorithm. Inspired by the above background, we have proposed a multi-agent MAB-based scheme to resolve joint load balancing and resource allocation in O-RAN. The major contributions of this paper are listed as follows.
\begin{itemize}
	\item We propose a novel mobility-aware load balancing and resource allocation (mmLBRA) scheme for O-RAN, which takes into account the dynamic nature of users and co-channel interference, and is scalable for a large number of users. The scheme employs multi-agent multi-armed bandit (MA-MAB) learning to optimize handover parameters and resource allocation for users, and can be operated in both Non-RT RIC and Near-RT RIC.
	
	\item We formulate an optimization problem for MLB in O-RAN. The difference of loading among O-RUs is minimized in a long-term manner, while the system rate should be optimized in a short term, which are managed by Non-RT RIC and Near-RT RIC, respectively. 
	
	\item Our proposed mmLBRA scheme shows significant performance improvements over the baseline method in terms of load balancing and effective network sum-rate. The proposed scheme can effectively allocate network resources and improve the QoS for users in a scalable O-RAN system, which outperforms the baseline and existing methods in open literature.
\end{itemize}

The rest of the paper is organized as follows. Section \ref{SM} will introduce the system model and the problem formulation of load balancing and resource allocation. Section \ref{proposed_scheme} elaborates the proposed mmLBRA scheme and the default method in the specification. Section \ref{PE} provides the performance evaluation through simulations, whilst conclusions are drawn in Section \ref{CON}.

\section{System Model and Problem Formulation}\label{SM}

\begin{figure}
\centering
 \includegraphics[width=2.7in]{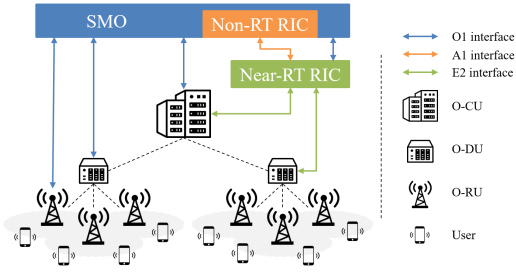}
 \caption{O-RAN architecture.} \label{oran_architecture}
\end{figure}

\subsection{Network Architecture and System Model}
The network architecture we examine, as illustrated in Fig. \ref{oran_architecture}, is based on the O-RAN design, consisting of multiple O-RAN network functions such as service management and orchestration (SMO), Near-RT RIC, O-CU, O-DU, and O-RU. These functions communicate with one another using the open interface of O-RAN. The Near-RT RIC is linked to E2 nodes (O-CUs, O-DUs) via the E2 interface, enabling the optimization of the system or allocation of resources. Non-RT RIC provides AI/ML management services to the Near-RT RIC via the A1 interface.

The network comprises one O-CU, multiple O-DUs $\mathcal{G} = \{1,..., G\}$, multiple O-RUs $\mathcal{S} = \{1,..., S\}$, and multiple serving users $\mathcal{U} = \{1,..., U\}$. Each user can only be served by one O-RU, while one O-DU can handle multiple O-RUs. An O-CU connects to multiple O-DUs and communicates with the core network via backhaul links. Each O-RU has $N$ available subchannels $\mathcal{N} = \{1, ..., N\}$, with a bandwidth of $W$ per subchannel. Only one subchannel can be assigned to a user at a time. Due to the sharing of the frequency band, interference exists between different O-RUs. The signal-to-interference-plus-noise ratio (SINR) of a user $u$ served by O-RU $s$ connected to O-DU $g$ on subchannel $n$ is given by
\begin{equation} \label{SINR}
\gamma_{g,s,u}^n = \frac{p_{g,s,u}^nh_{g,s,u}^n}{I_{g,s,u}^n+N_{0}W},
\end{equation}
where $p_{g,s,u}^n$ and $h_{g,s,u}^n$ denote the transmit power and channel gain of O-RU $s$ served by O-DU $g$ to user $u$ on subchannel $n$, respectively. $N_{0}$ is the power spectral density of Gaussian noise, and $W$ is the subchannel bandwidth. $I_{g,s,u}^n$ quantifies the co-channel interference that user $u$, served by O-RU $s$ connected to O-DU $g$ on subchannel $n$, experiences from other O-RUs in the network, which can be expressed as
\begin{align}
	I_{g,s,u}^n &= \sum_{\substack{k=1 \\ k \neq u}}^U \phi_{g,s,k}\psi_{g,s,k}^n p_{g,s,k}^n h_{g,s,u}^n
\notag\\
	&+ \sum_{\substack{j=1 \\ j \neq s}}^S\sum_{\substack{k=1 \\ k \neq u}}^U \phi_{g,j,k}\psi_{g,j,k}^n p_{g,j,k}^n h_{g,j,u}^n
\notag\\
	&+ \sum_{\substack{i=1 \\ i \neq g}}^G\sum_{\substack{j=1 \\ j \neq s}}^S\sum_{\substack{k=1 \\ k \neq u}}^U \phi_{i,j,k}\psi_{i,j,k}^n p_{i,j,k}^n h_{i,j,u}^n,
\end{align}
where $\phi_{i,j,k}\in \{0,1\}$ indicates whether user $k$ is associated with O-RU $j$ connected to O-DU $i$. $\psi_{i,j,k}^n \in \{0,1\}$ denotes the decision to allocate subchannel $n$ to user $k$ served by O-RU $j$ connected to O-DU $i$. Accordingly, the candidate resource allocation (RA) solution set of subchannel allocation (SA) is $\Psi = \{\psi_{g,s,u}^n\big|1\leq g \leq G, 1\leq s \leq S, 1\leq u \leq U, 1\leq n \leq N\}$, whereas power allocation (PA) is $\mathrm{P} = \{p_{g,s,u}^n\big|1\leq g \leq G, 1\leq s \leq S, 1\leq u \leq U, 1\leq n \leq N\}$. Moreover, O-RUs can allocate different numbers of subchannels to users based on their data rate requirements. The number of subchannels obtained by user $u$ served by O-RU $s$ connected to O-DU $g$ can be calculated by
\begin{equation}
\Omega_{g,s,u} = \sum_{n=1}^N \psi_{g,s,u}^n = \bigg \lceil \frac{D_{g,s,u}}{C_{g,s}W} \bigg \rceil,
\end{equation}
where $D_{g,u,s}$ is the user $u$ data rate demand served by O-RU $s$ connected to O-DU $g$. $C_{g,s}$ is the spectrum efficiency of O-RU $s$ connected to O-DU $g$. Therefore, the subchannel utilization of O-RU $s$ connected to O-DU $g$ can be calculated as
\begin{equation}
\Omega_{g,s} = \sum_{u=1}^U \phi_{g,s,u}\sum_{n=1}^N \psi_{g,s,u}^n,
\end{equation}
which represents the loading of the O-RU. Hence, the subchannel utilization difference between O-RU $s$ connected to O-DU $g$ and other O-RUs is calculated as
\begin{equation}
\eta_{g,s} = \sum_{\substack{i=1 \\ i \neq g}}^G\sum_{\substack{j=1 \\ j \neq s}}^S \chi_{i,j}\big|\Omega_{g,s} - \Omega_{i,j}\big|,
\end{equation}
where $\chi_{i,j}\in \{0,1\}$ indicates whether O-RU $j$ is connected to O-DU $i$, where the set is denoted as $\Xi=\{ \chi_{i,j}| 1\leq i\leq G, 1\leq j\leq S \}$. Given SINR in $\eqref{SINR}$, the achievable data rate of user $u$ served by O-RU $s$ connected to O-DU $g$ can be formulated based on Shannon capacity as
\begin{equation}
R_{g,s,u} = \sum_{n=1}^N \psi_{g,s,u}^nW\log_{2}(1+\gamma_{g,s,u}^n).
\end{equation}
Thus, the sum-rate of O-RU $s$ connected to O-DU $g$ can be calculated by
\begin{equation}
\label{sum-rate}
R_{g,s} = \sum_{u=1}^U \phi_{g,s,u}R_{g,s,u}.
\end{equation}

\subsection{Handover Mechanism}
\begin{figure}
\centering
 \includegraphics[width=2.7in]{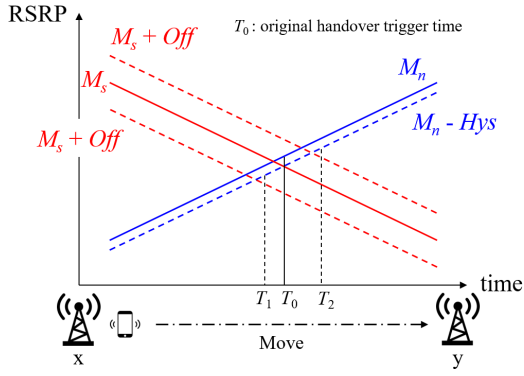}
 \caption{The RSRP of user when moving from serving O-RU \emph{x} to neighboring O-RU \emph{y}. If the $Off<0$ and $Hys>0$, the handover trigger time will change from $T_{0}$ to $T_{1}$. If $Off>0$ and $Hys>0$, the handover trigger time will shift from $T_{0}$ to $T_{2}$.} \label{A3Event}
\end{figure}

As defined in the specification, we adopt the A3 event to determine whether the user is ready to be handed over, which is defined as follows:
\begin{equation} \label{TrigCon}
\text{Trigger condition: } M_n - Hys > M_s + Off
\end{equation}
\begin{equation} \label{LeavCon}
\text{Leave condition: } M_n + Hys < M_s + Off
\end{equation}
Here, $M_n$ and $M_s$ are the reference signal received power (RSRP) of the neighboring O-RUs and serving O-RU, respectively. $Off$ and $Hys$ are handover parameters, where the former is the A3 event offset while the latter is hysteresis. The trigger condition in $\eqref{TrigCon}$ is satisfied when the RSRP of the neighboring O-RUs is greater than that of the serving O-RU by an offset. If the RSRP of the user satisfies $\eqref{LeavCon}$ within the time-to-trigger (TTT) phase, the user will continue to be served by the original O-RU; otherwise, the user will be offloaded to an appropriate target O-RU. The set of available handover parameter pairs for O-RUs can be defined as $\Theta = \{(Off, Hys):Off_{min} \leq Off \leq Off_{max}, Hys_{min} \leq Hys \leq Hys_{max}\}$. As shown in Fig. \ref{A3Event}, we can accelerate or delay the handovers of users by changing the values of both parameters in O-RUs to attain load balancing of the network. Therefore, the association between O-RUs and users can be expressed as $\Phi = \{\phi_{g,s,u}\big|1\leq g \leq G, 1\leq s \leq S, 1\leq u \leq U\}$, where $\Lambda = \{\lambda_{g,s} \in \Theta \big| 1\leq g \leq G, 1\leq s \leq S\}$ is the set of handover parameter pairs for each O-RU.

\subsection{Problem Formulation}
We formulate a problem for joint MLB and RA, with the objective utility function of minimizing the difference in subchannel utilization between O-RUs while maximizing the individual throughput. This problem takes into account the adjustment of handover parameters of O-RUs of $\{\Xi,\Phi,\Lambda\}$, subchannel allocation $\Psi$, and power allocation $\mathrm{P}$. The problem can be expressed as
\begingroup
\allowdisplaybreaks
\begin{subequations} \label{pro1}
\begin{align}
	&\min_{\Xi,\Phi,\Lambda,\Psi,\mathrm{P}} && \sum_{g = 1}^G\sum_{s = 1}^S \chi_{g,s} \left[ \eta_{g,s}- \kappa_{g,s} (1-p_{o}) R_{g,s}\right]  \label{Obj1} \\
	&\qquad\text{ s.t.} &&\chi_{g,s}, \phi_{g,s,u}, \psi_{g,s,u}^n \in \{0,1\}, \hspace{0.3cm} \forall g, \forall s, \forall u, \forall n, \label{C1} \\
	&&& \sum_{s = 1}^S \phi_{g,s,u} \leq 1, \hspace{2.1cm} \forall g, \forall u, \label{C2} \\
	&&& \sum_{g = 1}^G \chi_{g,s} = 1, \hspace{2.3cm} \forall s, \label{C3} \\
	&&& \sum_{u=1}^{U}\sum_{n=1}^{N} P_{g,s,u}^{n} \leq P_{g,s}^{(\mathrm{Max})}, \hspace{0.58cm} \forall g, \forall s, \label{C4} \\
	&&& \sum_{u=1}^U \phi_{g,s,u}R_{g,s,u} \leq C_{g,s}^{(\mathrm{Max})}, \hspace{0.3cm} \forall g, \forall s, \label{C5} \\
	&&& \lambda_{g,s} \in \Theta, \hspace{2.9cm} \forall g, \forall s. \label{C6}
\end{align}
\end{subequations}
\endgroup
In $\eqref{Obj1}$, $\kappa_{g,s}\geq 0$ is defined as a positive penalty term reflecting the importance of either load balancing or throughput performance. Also, $p_o$ indicates the averaged outage probability. Constraint $\eqref{C1}$ represents the binary set constraint which specifies that each link connection can be either on or off. Constraint $\eqref{C2}$ indicates that each user can be served by only one O-RU. Constraint $\eqref{C3}$ ensures that each O-RU is connected to only one O-DU. Constraint $\eqref{C4}$ restricts the transmit power of each O-RU. Constraint $\eqref{C5}$ ensures that the sum-rate of each O-RU is less than or equal to its corresponding fronthaul capacity allowance $C_{g,s}^{(\mathrm{Max})}$. Finally, constraint $\eqref{C6}$ expresses that the handover parameter pairs chosen by each O-RU must be available. The problem is NP-hard due to coupled parameters and non-convexity and non-linearity of subchannel utilization and throughput functions, which is addressed in the following section.

\section{Proposed muti-agent MAB for load balancing and resource allocation (mmLBRA) scheme}\label{proposed_scheme}

\begin{figure}
\centering
\includegraphics[width=2.8in]{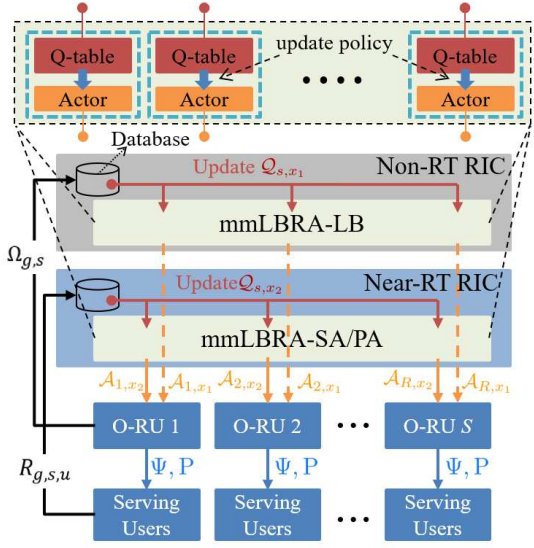}
\caption{The proposed mmLBRA scheme in O-RAN. mmLBRA-LB is conducted in Non-RT RIC as rApps, whilst mmLBRA-SA/PA is performed in Near-RT RIC as xApps.} \label{RL_framework}
\end{figure}

Conventionally, a BS can centrally solve the joint optimization problem in $\eqref{pro1}$, but the computational complexity is unprecedentedly high. Alternatively, under the architecture of O-RAN, functions can be split into RICs that perform long-term and short-term determinations, resulting in a beneficial reduction in computational loads. Therefore, we partition the problem by decomposing $\eqref{pro1}$ into two subproblems. In the Non-RT RIC with rApps, the LB subproblem is solved, which minimizes the subchannel utilization difference between O-RUs by adjusting association and handover-related parameters, which is demonstrated as
\begin{subequations} \label{pro2}
\begin{align}
	&\min_{\Xi,\Phi,\Lambda} && \sum_{g = 1}^G\sum_{s = 1}^S \chi_{g,s}\eta_{g,s}(\Xi,\Phi,\Lambda \big|\Psi,\mathrm{P}) \label{Obj2} \\
	&\quad\text{ s.t.} && \chi_{g,s}, \phi_{g,s,u} \in \{0,1\}, \hspace{0.3cm} \forall g, \forall s, \forall u, \label{O2C1} \\
&&& \eqref{C2}, \eqref{C3}, \eqref{C5}, \eqref{C6}.
\end{align}
\end{subequations}
On the other hand, the aim of the remaining subproblem 2 is to maximize the effective sum-rate by executing SA and PA. Such problem should be performed in Near-RT RIC as xApps in a short-term period complied with the fluctuated channel coherence time, which is represented as
\begin{subequations} \label{pro3}
\begin{align}
	&\max_{\Psi,\mathrm{P}} && (1-p_{o})\sum_{g = 1}^G\sum_{s = 1}^S \chi_{g,s}R_{g,s}(\Psi,\mathrm{P}\big|\Xi,\Phi,\Lambda), \label{Obj3} \\
	&\quad\text{ s.t.} && \psi_{g,s,u}^n \in \{0,1\}, \hspace{0.3cm} \forall g, \forall s, \forall u, \forall n, \label{O3C1} \\
	&&& \eqref{C4}, \eqref{C5}.
\end{align}
\end{subequations}
We propose the mmLBRA scheme as a solution to solve the two subproblems. To achieve this, we adopt MAB, a low computing complexity method that can work in dynamic and uncertain environments. The learning framework of MAB includes an agent and action, and reward value, where the agent updates the policy by interacting with the environment to make intelligent decisions. Note that compared to RL method, we do not require higher-dimensional states for optimization. Fig. \ref{RL_framework} illustrates the proposed mmLBRA scheme. To avoid the large searching space of actions induced by centralized learning, we have designed a distributed synchronous learning architecture with the concept of multi-agent. The mmLBRA scheme includes two parts, mmLBRA-LB and mmLBRA-RA. Considering the control loop time of RIC, the former is performed in Non-RT RIC while the latter is implemented in Near-RT to allocate resources to users in real-time. Furthermore, mmLBRA-RA comprises mmLBRA-SA and mmLBRA-PA sub-schemes. Due to high-complexity of solving continuous solution of transmit power, we take design mmLBRA-PA by using a multi-level based mechanism. Therefore, in the following, we use the separate terms of mmLBRA-PA and mmLBRA-SA.

In proposed mmLBRA scheme, the O-RU $s$ acts as an agent that will take the action at each time $t$, $\mathcal{A}_{s,\mathcal{X}}^t$, and then receive the reward $\mathcal{R}_{s,\mathcal{X}}^t$ from the environment, where $\mathcal{X}\in\{x_{1},x_{2}\}$ indicates whether the definition belongs to mmLBRA-LB or mmLBRA-RA. Some essential definitions are given as follows. For mmLBRA-LB, we define $\mathcal{A}_{s,x_{1}}^t=\{\Xi_{g,s},\Phi_{g,s},\Lambda_{g,s}\}$, $\mathcal{R}_{s,x_{1}}^t=-\eta_{g,s}$. Note that negativity is introduced due to maximization of a reward in MAB. As for mmLBRA-PA, we define $\mathcal{A}_{s,x_{2}}^t=\{\Psi,\mathrm{P}\}$ as well as $\mathcal{R}_{s,x_{2}}^t = R_{g,s,u}$. The MAB update is expressed as:
\begin{equation}
\label{QvalueUpdate}
\mathcal{Q}_{s,\mathcal{X}}^t(\mathcal{A}_{s,\mathcal{X}}^t) = \mathcal{Q}_{s,\mathcal{X}}^{t-1}(\mathcal{A}_{s,\mathcal{X}}^t) + \frac{\mathcal{R}_{s,\mathcal{X}}^t - \mathcal{Q}_{s,\mathcal{X}}^{t-1}(\mathcal{A}_{s,\mathcal{X}}^t)}{N(\mathcal{A}_{s,\mathcal{X}}^t)},
\end{equation}
where $N(\mathcal{A}_{s,\mathcal{X}}^t)$ is the number of times action $\mathcal{A}_{s,\mathcal{X}}^t$ is selected by O-RU $s$ at time $t$. To attain optimal convergence, the agent must trade-off between selecting actions randomly (exploration) and choosing actions based on the value function (exploitation) with probability $\epsilon$. Unlike the $\epsilon$-greedy strategy, the Upper Confidence Bound (UCB) algorithm is adopted in mmLBRA to prevent frequent selection of a certain action. The corresponding action is designed as
\begin{subequations}
\begin{numcases}
{\mathcal{A}_{s,\mathcal{X}}^t=}
    \mathop{\arg\max}\limits_{\mathcal{A}_{s,\mathcal{X}}^{\prime}} \mathcal{Q}_{s,\mathcal{X}}^t(\mathcal{A}'_{s,\mathcal{X}}) + \sqrt{\frac{2\cdot \ln (m)}{N(\mathcal{A}'_{s,\mathcal{X}})}}, \label{policy1} \\
    \text{Randomly select untried action } \mathcal{A}_{s,\mathcal{X}}^{\prime}. \label{policy2}
\end{numcases}
\end{subequations}
If all actions $\mathcal{A}_{s,\mathcal{X}}$ have been tried, the agent will follow $\eqref{policy1}$ to select the action; otherwise, it will follow $\eqref{policy2}$. In $\eqref{policy1}$, $m = \sum_{a\in\mathcal{A}_{s,\mathcal{X}}} N(a)$, and when the action is selected frequently, the action-value will be reduced. Therefore, the agent has the opportunity to select other actions to prevent the locally-optimal solution and reach a balance between exploitation and exploration without setting the probability $\epsilon$. The overall algorithm of the proposed mmLBRA scheme is shown in Algorithm \ref{mmLBRA}, which requires mmLBRA-PA in Algorithm \ref{PA}.

\begin{algorithm}[!tb]
\footnotesize
  \caption{Proposed mmLBRA scheme}
  \SetAlgoLined
  \DontPrintSemicolon
  \label{mmLBRA}
  \begin{algorithmic}[1]
  \STATE Initialize: $iter,T_{1},T_{2},T_{3},T_{4},\Xi,\Phi,\Lambda,\Psi,\mathrm{P}$
  \STATE Set $\mathcal{Q}_{s,\mathcal{X}}(\mathcal{A}_{s,\mathcal{X}})=\emptyset,\forall\mathcal{X}=\{x_{1},x_{2}\}$
  \FOR{$t=1:iter$}
    \FOR{each O-RU $s$}
       \STATE \textbf{Non-RT RIC performs mmLBRA-LB ($T_{1}$ period):} 
       \\ $\mathcal{X} \leftarrow x_{1}$, conduct $\eqref{policy1}$ and $\eqref{policy2}$ for adjusting  association and handover parameters of $\Xi,\Phi,\Lambda$  
       \STATE \textbf{Near-RT RIC conducts mmLBRA-RA ($T_{3}$ period):}
       \\ $\mathcal{X} \leftarrow x_{2}$
       \\ 1) Perform mmLBRA-SA by executing $\eqref{policy1}$ and $\eqref{policy2}$ for allocating subchannel indicators $\Psi$
       \\ 2) Perform mmLBRA-PA sub-scheme in Algorithm \ref{PA} for allocating power   $\mathrm{P}$
    \ENDFOR
    \STATE Obtain candidate solution of $\Xi,\Phi,\Lambda,\Psi,\mathrm{P}$
    \FOR{each O-RU $s$}
    	\STATE Calculate the reward $\mathcal{R}_{s,\mathcal{X}}^{t}$, and update the action-value function $\mathcal{Q}_{s,\mathcal{X}}^{t}$ by $\eqref{QvalueUpdate}$ for \textbf{mmLBRA-LB in Non-RT RIC} (with $T_{2}$ period) and \textbf{mmLBRA-SA in Near-RT RIC} (with $T_{4}$ period)
    \ENDFOR
    \STATE Update locations and channel states of users
  \ENDFOR
  \end{algorithmic}
\end{algorithm}

\begin{algorithm}[!tb]
\footnotesize
  \caption{mmLBRA-PA sub-scheme}
  \SetAlgoLined
  \DontPrintSemicolon
  \label{PA}
  \begin{algorithmic}[1]
  \STATE Input: RSRP of user $u$
  \IF{$RSRP \geq RSRP_{T1}$}
      \STATE $p_{g,s,u}^n \leftarrow p_{4}$
  \ELSIF{$RSRP \geq RSRP_{T2}$}
      \STATE $p_{g,s,u}^n \leftarrow p_{3}$
  \ELSIF{$RSRP \geq RSRP_{T3}$}
      \STATE $p_{g,s,u}^n \leftarrow p_{2}$
  \ELSE
      \STATE $p_{g,s,u}^n \leftarrow p_{1}$
  \ENDIF
  \STATE Output: the transmit power of user $u$
  \end{algorithmic}
\end{algorithm}

\section{Performance Evaluation}\label{PE}


\begin{table}[t]
\centering
\scriptsize
  \caption{Parameters setting}
 \begin{tabular}{l*{1}{l}r}
  \hline
  $\textbf{System Parameter}$ & $\textbf{Value}$ \\
  \hline
  Number of O-DUs/O-RUs  & $\{3,7\}$    \\
  Number of subchannels & $80$      \\
  Subchannel bandwidth, $W$ & $360$ kHz \\
  Available transmit power, $\{p_{1},p_{2},p_{3},p_{4}\}$ & $\{0.5,1.25,3.75,5\}$ W \\
  Centre frequency, $f_{c}$ & $3.5$ GHz \\
  Antenna height at O-RU/UE & $\{25,1.5\}$ m  \\
  Noise power, $N_{0}$ & $-174$ dBm/Hz \\
  A3 event offset, $Off$ & $\{-15,-14,...,15\}$ dB\\
  Hysteresis, $Hys$ & $\{0,1,...,15\}$ dB \\
  Time to trigger, TTT & $2.56$ s \\
  Total duration of simulation time, $iter$ & $30000$ s \\
  Period of $\{T_{1},T_2,T_3,T_4\}$ & $\{10,1,1,1\}$ s \\
  $\{RSRP_{T1},RSRP_{T2},RSRP_{T3}\}$ & $\{-80,-90,-100\}$ dBm \\
  \hline
 \end{tabular} \label{Parameter}
\end{table}

We conducted a simulation of a typical hexagonal cellular network consisting of $7$ O-RUs, with an inter-site distance of $500$ m, and employed the 3GPP channel model for UMa scenario \cite{channelmodel}. The network included $80$ users, of which $70\%$ were restricted to a particular cell, causing that cell to become overloaded, while the remaining $30\%$ were randomly and uniformly distributed. The users' initial positions were randomly determined, and they moved in any direction at a constant speed. The simulation used the system parameters listed in Table \ref{Parameter}. We compared our proposed mmLBRA scheme with the following benchmarks.
\begin{itemize}
	\item \textbf{Default}: Handover parameters are default set as a fixed value of $0$ dB. Conventional A3-event offset is performed based on the conditions of $\eqref{TrigCon}$ and $\eqref{LeavCon}$ \cite{compare2}.
	\item \textbf{rLBRA}: Rule-based LBRA (rLBRA) method is compared as a benchmark. A3 event offset of O-RUs is determined based its loading. For instance, the offset and hysteresis are set to become proportional to the subchannel utilizations, i.e., loads of an O-RU, which aims for offloads those overloaded O-RUs. Note that conventional maximum throughput scheduler is applied in rLBRA \cite{compare1}.
	\item $\epsilon$-\textbf{greedy}: Unlike UCB in mmLBRA, it \cite{compare3} only selects an appropriate action under a fixed exploration probability, i.e., number of actions is not considered.
	\item \textbf{Without RA}: Conventional round-robin scheduler and equal power control are adopted. Handover parameters are determined by mmLBRA-LB.
\end{itemize}

\begin{figure}
\centering
\includegraphics[width=2.7in]{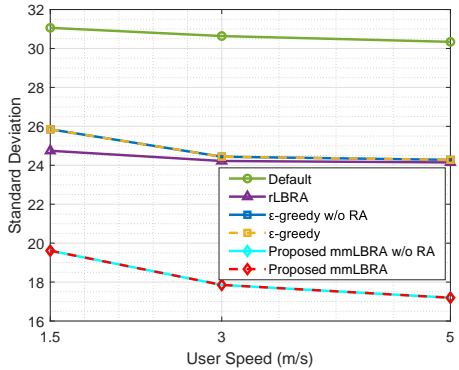}
\caption{Performance of standard deviation of the subchannel utilization among O-RUs under different user speeds.} \label{std1}
\end{figure}

We have investigated the impact of user speed on network load balancing using different algorithms and observed the standard deviation of the O-RU subchannel utilization, which reflects the uniformity of loads among O-RUs. As shown in Fig. \ref{std1}, the LB method had a smaller standard deviation value, regardless of the user speed, while the method without LB had the largest deviation. This suggests that adjusting the handover parameter through the algorithm can achieve load balancing in the network when users are moving, with the proposed scheme having the best performance. Moreover, faster user movement resulted in a more balanced load distribution among O-RUs. This is because faster user speeds increased the likelihood of offloading, thereby reducing the burden on heavily loaded O-RUs and increasing the load on lightly loaded O-RUs, leading to a slightly balanced load distribution. Conversely, slower user speeds led to lower handover frequencies, causing heavily loaded O-RUs to remain busy for longer periods while lightly loaded O-RUs remained idle, exacerbating the load imbalance in the network.

\begin{figure}
\centering
\includegraphics[width=2.7in]{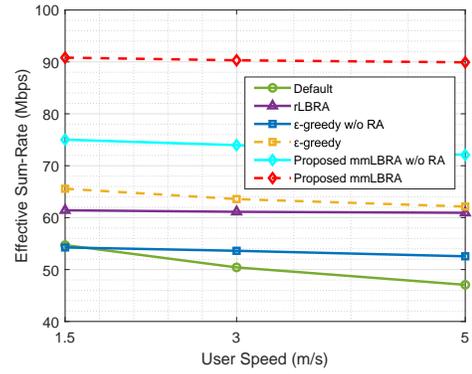}
\caption{Performance of effective sum-rate under different user speeds.} \label{rate1}
\end{figure}

In this study, we also examined the effect of user speed on the effective sum-rate, as shown in Fig. \ref{rate1}. The default method connects users to the O-RU with the strongest signal strength, resulting in a higher sum-rate but no LB. However, this leads to a much higher outage probability for users than the methods with LB. Consequently, the network has a lower effective sum-rate without LB, whereas the effective sum-rate significantly improves with the methods with LB, especially with the proposed mmLBRA scheme, which outperforms the rLBRA and $\epsilon$-greedy methods. Additionally, the addition of RA to the proposed scheme further increases the effective sum-rate by about $20$ Mbps. However, when the user speed increases, the effective sum-rate slightly decreases. This is because slower-moving users can occupy the channels for a longer duration with less frequent channel state changes, enabling better QoS. On the other hand, faster-moving users experience frequent channel state changes and are more likely to be offloaded, making it difficult to maintain QoS and resulting in a decline in effective sum-rate.

\begin{figure}
\centering
\includegraphics[width=2.7in]{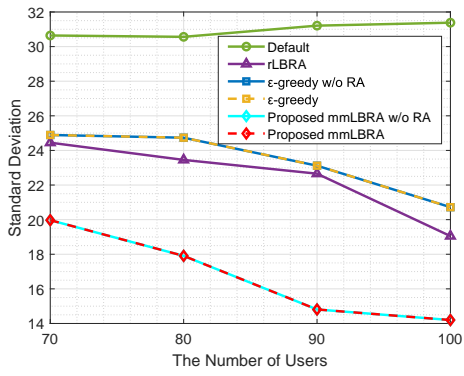}
\caption{Performance of standard deviation under different numebrs of users.} \label{std}
\end{figure}

In Fig. \ref{std}, we investigated the impact of the number of users on network load balancing and effective sum-rate for users with equal data-rate QoS and moving speed. The results are presented in Fig. \ref{std} where the method with LB, especially the proposed mmLBRA scheme, outperforms the method without LB, which is the worst. In the default method, most users are connected to overloaded O-RU due to its strong signal strength, resulting in an imbalanced network with that O-RU being overloaded while the other O-RUs are idle. However, the method with LB can adjust the handover parameter and offload some users from the overloaded O-RU to other O-RUs, reducing the loading on that O-RU and increasing the loading on other O-RUs. Consequently, the loads among O-RUs become more balanced, and more users are active. Moreover, as the number of users increases in the default method, the standard deviation slightly increases but not significantly. Conversely, the standard deviation decreases as the number of users increases when using the method with LB. When more users are connected to the originally overloaded O-RU, the method with LB offloads more users to achieve load balancing. This causes the loads of other O-RUs to increase, while the loading difference among O-RUs becomes smaller, leading to a more balanced network.

\begin{figure}
\centering
\includegraphics[width=2.7in]{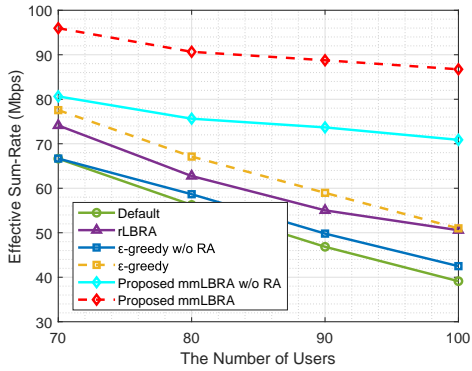}
\caption{Performance of effective sum-rate under different numbers of users.} \label{rate}
\end{figure}

In Fig. \ref{rate}, we observe the impact of the number of users on the effective sum-rate. The method with LB yields a higher effective sum-rate compared to the default method, which has the lowest effective sum-rate. This is because, although the default method has a higher sum-rate, the user outage probability is much larger than that of the method with LB, leading to a significant reduction in the effective sum-rate. For the LB method, some users in the middle of the cell are offloaded early after adjusting the handover parameter of O-RUs, leading to a decrease in the sum-rate. However, the outage users at the edge of the cell can obtain the subchannels after being offloaded, resulting in a lower user outage probability and a better performance of the effective sum-rate. Moreover, as the number of users increases, more users share the entire frequency resource, leading to an increase in the sum-rate of the network. However, due to the limited number of available subchannels, the user outage probability is higher than in the case of fewer users, resulting in a lower effective sum-rate.

\section{Conclusion}\label{CON}
In this paper, we have proposed the mmLBRA scheme to tackle the issue of joint load imbalance and resource allocation among O-DUs/RUs in O-RAN. The mmLBRA scheme consists of two sub-schemes, namely mmLBRA-LB and mmLBRA-RA, which can be executed separately on rApp in Non-RT RIC and xApp in Near-RT RIC platforms. Leveraging the multi-agent MAB and UCB algorithms as the mechanism for policy updating, the proposed scheme enables the O-RUs to learn the optimal handover parameter settings and resource allocations, while avoiding local optima. The results demonstrate that our proposed mmLBRA scheme outperforms the baseline and existing benchmarks in terms of moderate loads and higher effective sum-rate under non-uniform user distribution.

\linespread{0.9} 
\scriptsize
\bibliographystyle{IEEEtran}
\bibliography{IEEEabrv}
\end{document}